\documentclass[journal,transmag]{IEEEtran}
\ifCLASSINFOpdf
\else
\fi
\usepackage{times}
\usepackage{epsfig}
\usepackage{graphicx}
\usepackage{amsmath}
\usepackage{amssymb}
\usepackage{subfigure}
\usepackage{multirow}
\usepackage{epstopdf}
\usepackage{color}

\usepackage{paralist}
\let\itemize\compactitem
\let\enditemize\endcompactitem
\let\enumerate\compactenum
\let\endenumerate\endcompactenum

\begin{document}
%
\title{A Light CNN for Deep Face Representation with Noisy Labels}


\author{\IEEEauthorblockN{Xiang Wu,
Ran He,~\IEEEmembership{Senior Member,~IEEE},
Zhenan Sun$^*$,~\IEEEmembership{Member,~IEEE}, and
Tieniu Tan,~\IEEEmembership{Fellow,~IEEE}}
\thanks{
$^*$Zhenan~Sun is the corresponding author. X. Wu, R. He, Z. Sun and T. Tan are with National Laboratory of Pattern Recognition, CASIA, Center for Research on Intelligent Perception and Computing, CASIA, Center for Excellence in Brain Science and Intelligence Technology, CAS and University of Chinese Academy of Sciences, Beijing, China,100190.

E-mail:alfredxiangwu@gmail.com, \{rhe, znsun, tnt\}@nlpr.ia.ac.cn}}

\markboth{Journal of \LaTeX\ Class Files,~Vol.~14, No.~8, August~2017}%
{Shell \MakeLowercase{\textit{et al.}}: Bare Demo of IEEEtran.cls for IEEE Transactions on Magnetics Journals}
%



\IEEEtitleabstractindextext{%
\begin{abstract}
The volume of convolutional neural network (CNN) models proposed for face recognition has been continuously growing larger to better fit large amount of training data. When training data are obtained from internet, the labels are likely to be ambiguous and inaccurate. This paper presents a Light CNN framework to learn a compact embedding on the large-scale face data with massive noisy labels. First, we introduce a variation of maxout activation, called Max-Feature-Map (MFM), into each convolutional layer of CNN. Different from maxout activation that uses many feature maps to linearly approximate an arbitrary convex activation function, MFM does so via a competitive relationship. MFM can not only separate noisy and informative signals but also play the role of feature selection between two feature maps. Second, three networks are carefully designed to obtain better performance meanwhile reducing the number of parameters and computational costs. Lastly, a semantic bootstrapping method is proposed to make the prediction of the networks more consistent with noisy labels. Experimental results show that the proposed framework can utilize large-scale noisy data to learn a Light model that is efficient in computational costs and storage spaces. The learned single network with a 256-D representation achieves state-of-the-art results on various face benchmarks without fine-tuning. The code is released on https://github.com/AlfredXiangWu/LightCNN.
\end{abstract}

}

\maketitle

\IEEEdisplaynontitleabstractindextext

%
\IEEEpeerreviewmaketitle

\section{Introduction}

In the last decade, convolutional neural network (CNN) has become one of the most popular techniques for solving computer vision problems. Numerous vision tasks, such as image classification \cite{HeZRS16}, object detection \cite{DBLP:journals/corr/RedmonDGF15}, face recognition \cite{DBLP:conf/icb/LeiCHLL07, sun2014deep, taigman2014deepface, yi2014learning}, have benefited from the robust and discriminative representation learned via CNNs. As a result, their performances have been greatly improved, for example, the accuracy on the challenging labeled faces in the wild (LFW) benchmark has been improved from 97\% \cite{taigman2014deepface} to 99\% \cite{parkhi2015deep, schroff2015facenet, sun2014deep}. This improvement mainly benefits from which CNN can learn robust face embedding from the training data with lots of subjects. To achieve optimal accuracy, the scale of the training dataset for CNN has been consistently increasing. Some large-scale face datasets have been released such as CASIA-WebFace \cite{yi2014learning}, CelebFaces+ \cite{sun2014deep}, VGG face dataset \cite{parkhi2015deep}, UMDFace \cite{bansal2016umdfaces,bansal2017s}, MS-Celeb-1M \cite{DBLP:journals/corr/GuoZHHG16} and VGGFace2 dataset~\cite{DBLP:journals/corr/abs-1710-08092}. However, these large-scale datasets often contain massive noisy signals especially when they are automatically collected via image search engines or from movies.

\begin{figure}
\centering
\subfigure[ReLU: $h$($x$)=max(0, $x^1$)]{\centering \includegraphics[width=0.23\textwidth]{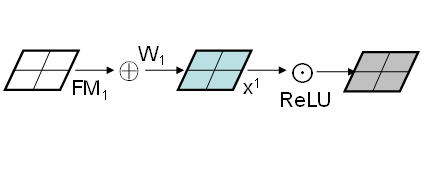}\label{fig:relu}}\hspace{0.2cm}
\subfigure[Maxout: $h$($x$)=max($x^i$)]{\centering \includegraphics[width=0.23\textwidth]{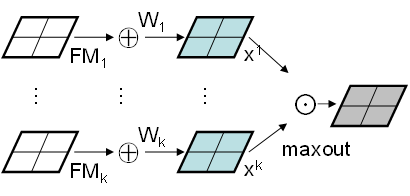}\label{fig:maxout}}
\subfigure[MFM 2/1: $h$($x$)=max($x^1$, $x^2$)]{\centering \includegraphics[width=0.23\textwidth]{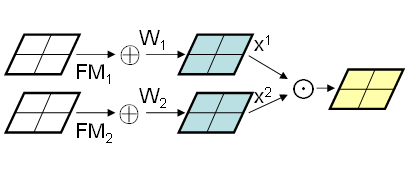}\label{fig:mfm_21}}
\subfigure[MFM 3/2: $h^1$($x$)=max($x^i$), $h^2$($x$)=median($x^i$)]{\centering \includegraphics[width=0.23\textwidth]{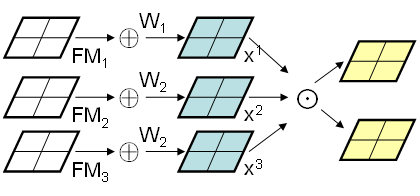}\label{fig:mfm_32}}
\caption{A comparison of different types of neural inhibition. (a) ReLU suppresses a neuron by thresholding magnitude responses. (b) Maxout with enough hidden units makes a piecewise linear approximation to an arbitrary convex function. (c) MFM 2/1 suppresses a neuron by a competitve relationship. It is the simplest case of maxout activations. (d) MFM 3/2 activates two neurons and suppresses one neuron.}
\label{fig:activation}
\end{figure}

This paper studies a Light CNN framework to learn a deep face representation from the large-scale data with massive noisy labels. As shown in Fig.~\ref{fig:activation}, we define a Max-Feature-Map (MFM) operation to obtain a compact representation and perform feature filter selection. MFM is an alternative of ReLU to suppress low-activation neurons in each layer, so that it can be considered as a special implementation of maxout activation \cite{goodfellow2013maxout} to separate noisy signals and informative signals. Our Light CNN architecture includes MFM, small convolution filters and Network in Network, and is trained on the MS-Celeb-1M dataset. To handle noisy labeled images, we propose a semantic bootstrapping method to automatically re-label training data via pre-trained deep networks. We assume that by given similar percepts, the model is capable of giving same predictions consistently.
Intuitively, too much skepticism of the original training label may lead to a wrong relabeling. Hence, it is important to balance the trade-off between the prediction and original labels. Extensive experimental evaluations demonstrate that the proposed Light CNN achieves state-of-the-art results on five face benchmarks without supervised fine-tuning. The contributions are summarized as follows:
\begin{itemize}\setlength{\itemsep}{1pt}
\item[1)] This paper introduces MFM operation, a special case of maxout to learn a Light CNN with a small number of parameters. Compared to ReLU whose threshold is learned from training data, MFM adopts a competitive relationship so that it has better generalization ability and is applicable on different data distributions.
\item[2)] The Light CNNs based on MFM are designed to learn a universal face representation. We propose three Light CNN models following the ideas of AlexNet, VGG and ResNet, respectively. The proposed models lead to better performance in terms of speed and storage space.
\item[3)] A semantic bootstrapping method via a pretrained deep network is proposed to handle noisy labeled images in a large-scale dataset. Inconsistent labels can be effectively detected by the probabilities of predictions, and then relabeled or removed for training.
\item[4)] The proposed single model with a 256-D representation obtains state-of-the-art results on various face benchmarks, i.e., large-scale, video-based, cross-age face recognition, heterogenous and cross-view face recognition datasets. The models contain fewer parameters and extract a face representation faster than other open source face models.
\end{itemize}

The paper is organized as follows. In Section II, we briefly review some related work on face recognition and noisy label problems. Section III describes the proposed Light CNN framework and the semantic bootstrapping method. Finally, we present experimental results in Section IV and conclude this paper in Section V.

\section{Related Work}
\subsection{CNN based Face Recognition}
Modern face recognition methods often regard CNNs as robust feature extractors.
Earlier DeepFace \cite{taigman2014deepface} trains CNN on 4.4M face images and uses CNN as a feature extractor for face verification.
It achieves 97.35\% accuracy on LFW with a 4096-D feature vector.
As an extension of DeepFace, Taigman \emph{et al.} \cite{taigman2014web} apply a semantic bootstrapping method to select an efficient training set from a large dataset.
Besides, it also discusses more stable protocols \cite{best2014unconstrained} of LFW, indicating the robustness of face features more representatively.
To further improve the accuracy, Sun \emph{et al.} \cite{sun2014deep} resorts to a multi-patch ensemble model.
An ensemble of 25 CNN models is trained on different local patches and Joint Bayesian is applied to obtain a robust embedding space.
In \cite{DBLP:conf/cvpr/SunWT15}, verification loss and classification loss are further combined to increase inter-class distance and decrease intra-class distance.
The ensemble model obtains 99.47\% on LFW.

Then, triplet loss is introduced into face recognition by FaceNet~\cite{schroff2015facenet}.
FaceNet is trained on about 100-200M face images with 8M face identities in total.
Since the selection of triplet pairs is important to achieve satisfying accuracy, FaceNet presents an online triplet mining method for training triplet-based CNN and achieves good accuracy (99.63\%).
Parkhi \emph{et al.} \cite{parkhi2015deep} train a VGG network \cite{simonyan2014very} on 2622 identities of 2.6M images collected from Internet and then fine-tune the model by a triplet-based metric learning method like FaceNet, which the accuracy achieves 98.95\% on LFW. Tran \emph{et al.} \cite{masi2016we} also propose a domain specific data augmentation to increase training data and obtains comparable performance on LFW.

The performance improvement of face recognition mainly benefits from CNN and large-scale face datasets. However, large-scale datasets often contain massive noisy labels especially when they are automatically collected from internet. Therefore, learning a Light CNN from the large-scale face data with massive noisy labels is of great significance.

\subsection{Noisy Label Problems}

Noisy label is an important issue in machine learning when datasets tend to be large. Some methods \cite{DBLP:journals/tnn/FrenayV14} have been devoted to deal with noisy label problems. These methods can generally be classified into three categories. In the first category, robust loss \cite{DBLP:conf/acl/BeigmanK09} is designed for classification tasks, which means the learned classification models are robust to the presence of label noise.  The second category \cite{DBLP:conf/icml/WilsonM97} aims to improve the quality of training data by identifying mislabeled instances. The third category \cite{DBLP:conf/icml/LawrenceS01} directly models the distribution of noisy labels during learning. The advantage of this category is that it allows using information about noisy labels during learning.

Recently, learning with noisy labeled data also draws much attention in deep learning, because deep learning is a data-driven approach and accurate label annotation is quite expensive. Mnih and Hinton \cite{DBLP:conf/icml/MnihH12} introduce two robust loss functions for noisy label aerial images. However, their method is only applicable for binary classification.
Sukhbaatar \emph{et al.} \cite{DBLP:journals/corr/SukhbaatarF14} consider multi-class classification for modeling class dependent noise distribution. They propose a bottom-up noise model to change the label probabilities output for back-propagation and a top-down model to do so given noisy labels before feeding data.
Moreover, with the notion of perceptual consistency, the work of \cite{DBLP:journals/corr/ReedLASER14} extends the softmax loss function by weakly supervised training. The idea is to dynamically update the targets of the prediction objective function based on the current model. They use a simple convex combination of training labels and the predictions of a current model to generate training targets. Although some strategies have been studied for noisy label problem, massive noisy label is still an open issue for deep learning methods.

\section{Architecture}

In this section, we first propose Max-Feature-Map operation for CNN to simulate neural inhibition, resulting in a new Light CNN framework for face analysis and recognition. Then, a semantic bootstrapping method for noisy labeled training dataset is addressed in detail.

\subsection{Max-Feature-Map Operation}
A large-scale face dataset often contains noisy signals. If the errors incurred by these noisy signals are not properly handled, CNN will learn a biased result. Rectified Linear Unit (ReLU) \cite{nair2010rectified} activation separates noisy signals and informative signals by a threshold (or bias) to determine the activation of one neuron. If the neuron is not active, its output value will be 0. However, this threshold might lead to the loss of some information especially for the first several convolution layers, because these layers are similar to Gabor filters (i.e., both positive and negative responses are respected). To alleviate this problem, Leaky Rectified Linear Units (LReLU) \cite{maas2013rectifier}, Parametric Rectified Linear Units (PReLU) \cite{DBLP:conf/iccv/HeZRS15} and Exponential Linear Units (ELU) \cite{DBLP:journals/corr/ClevertUH15} have been proposed.

In neural science, lateral inhibition (LI)~\cite{Amari1977DynamicsOP} increases the contrast and sharpness in visual or audio response and aids the mammalian brain in perceiving contrast within an image. Taking visual LI as an example, if an excitatory neural signal is released to horizontal cells, the horizontal cells will send an inhibitory signal to its neighboring or related cells. This inhibition produced by horizontal cells creates a more concentrated and balanced signal to cerebral cortex. As for auditory LI, if certain sound frequencies create a greater contribute to inhibition than excitation, tinnitus can be suppressed. Considering LI and noisy signals, we expect the activation function in one convolution layer to have the following characteristics:
\begin{itemize}\setlength{\itemsep}{1pt}
\item[1)] Since large-scale dataset often contains various types of noise, we expect that noisy signals and informative signals can be separated.
\item[2)] When there is a horizontal edge or line in an image, the neuron corresponding to horizontal information is excited whereas the neuron corresponding to vertical information is inhibited.
\item[3)] The inhibition of one neuron is free of parameters so that it does not depend on training data extensively.
\end{itemize}

To achieve the above characteristics, we propose the Max-Feature-Map (MFM) operation, which is an extension of Maxout activation \cite{goodfellow2013maxout}. However, the basic motivation of MFM and Maxout are different. Maxout aims to approximate an arbitrary convex function via enough hidden neurons. More neurons are used, better approximation results are obtained. Generally, the scale of a Maxout network is larger than that of a ReLU network. MFM resorts to max function to suppress the activations of a small number of neurons so that MFM based CNN models are light and robust. Although MFM and Maxout all use a max function for neuron activation, MFM cannot be treated as a convex function approximation. We define two types of MFM operations to obtain competitive feature maps.

Given an input convolution layer $x^n \in \mathbb{R}^{H\times W}$, where $n=\{1,...,2N\}$, $W$ and $H$ denote the spatial width and height of the feature map. The MFM 2/1 operation which combines two feature maps and outputs element-wise maximum one as shown in Fig.~\ref{fig:mfm_21} can be written as
\begin{equation}
\hat{x}^{k}_{ij}=\max(x^{k}_{ij}, x^{k+N}_{ij})
\label{eq:MFM}
\end{equation}
where the channel of the input convolution layer is $2N$, $1 \leq k \leq N, 1\leq i \leq H$, $1\leq j\leq W$. As is shown in Eq.~(\ref{eq:MFM}), the output $\hat{x}$ via MFM operation is in $\mathbb{R}^{H\times W \times N}$.

The gradient of Eq.~(\ref{eq:MFM}) takes the following form,
\begin{equation}
\begin{array}{cl}
\frac{\partial{\hat{x}^k_{ij}}}{\partial{x^{k}_{ij}}} &= \left\{
    \begin{array}{l}
    1, \text{ if } x^{k}_{ij}\geq x^{k+N}_{ij}\\
    0, \text{ otherwise}
    \end{array}
\right. \\
\\
\frac{\partial{\hat{x}^k_{ij}}}{\partial{x^{k+N}_{ij}}} &= \left\{
    \begin{array}{l}
    0, \text{ if } x^{k}_{ij}\geq x^{k+N}_{ij}\\
    1, \text{ otherwise}
    \end{array}
\right.
\end{array}
\label{eq:MFM_gradient}
\end{equation}
By using MFM 2/1, we generally obtain 50\% informative neurons from input feature maps via the element-wise maximum operation across feature channels.

Furthermore, as shown in Fig.~\ref{fig:mfm_32}, to obtain more comparable feature maps, the MFM 3/2 operation, which inputs three feature maps and removes the minimal one element-wise, can be defined as:
\begin{equation}
\left\{
\begin{array}{l}
\hat{x}^{k_1}_{ij}=\text{max}(x^{k}_{ij}, x^{k+N}_{ij}, x^{k+2N}_{ij}) \\
\\
\hat{x}^{k_2}_{ij}=\text{median}(x^{k}_{ij}, x^{k+N}_{ij}, x^{k+2N}_{ij})
\end{array}
\right.
\label{eq:MFM_32}
\end{equation}
where $x^n \in \mathbb{R}^{H\times W}$,  $1 \leq n \leq 3N, 1 \leq k \leq N$ and median($\cdot$) is the median value of input feature maps. The gradient of MFM 3/2 is similar to Eq.~(\ref{eq:MFM_gradient}), in which the value of gradient is 1 when the feature map $x^{k}_{ij}$ is activated, and it is set to be 0 otherwise. In this way, we select and reserve 2/3 information from input feature maps.

\subsection{The Light CNN Framework}
In the context of CNN, MFM operation plays a similar role to local feature selection in biometrics.
MFM selects the optimal feature at each location learned by different filters. It results in binary gradient (1 and 0) to excite or suppress one neuron during back propagation. The binary gradient plays a similar role of famous ordinal measure \cite{DBLP:journals/pami/SunT09} which is widely used in biometrics.

A CNN with MFM can obtain a compact representation while the gradients of MFM layers are sparse. Due to the sparse gradient of MFM, on the one hand, when doing back propagation in training stage, stochastic gradient descent (SGD) can only make effects on the neuron of response variables; on the other hand, when extracting features for testing, MFM can obtain more competitive nodes from previous convolution layers by activating the maximum of two feature maps. These observations demonstrate the valuable properties of MFM, i.e., MFM could perform feature selection and facilitate to generate sparse connections.

In this section, we discuss three architectures for our Light CNN framework. The first one is constructed by 4 convolution layers with Max-Feature-Map operations and 4 max-pooling layers like Alexnet \cite{DBLP:conf/nips/KrizhevskySH12} (as shown in Table \ref{tab:network_4layer}). It contains about 4,095K parameters and 1.5G FLOPS.

\begin{table}[h]
\centering
\caption{The architectures of the Light CNN-4 model.}
\begin{tabular}{|c|c|c|c|}
\hline
Type &\begin{tabular}{c}Filter Size\\/Stride\end{tabular} & Output Size & \#Params \\
\hline
Conv1 & $9 \times 9/1$ & $120\times 120 \times 96$ & 7.7K\\
MFM1 & - & $120\times 120 \times 48$ & - \\
\hline
Pool1 & $2 \times 2/2$ & $60\times 60 \times 48$ & -\\
\hline
Conv2 & $5 \times 5/1$ & $56\times 56 \times 192$ & 230.4K\\
MFM2 & - & $56\times 56 \times 96$ & - \\
\hline
Pool2 & $2 \times 2/2$ & $28\times 28 \times 96$ & -\\
\hline
Conv3 & $5 \times 5/1$ & $24\times 24 \times 256$ & 614K\\
MFM3 & - & $24\times 24 \times 128$ & - \\
\hline
Pool3 & $2 \times 2/2$ & $12\times 12 \times 128$ & -\\
\hline
Conv4 & $4 \times 4/1$ & $9\times 9 \times 384$ & 786K\\
MFM4 & - & $9\times 9 \times 192$ & - \\
\hline
Pool4 & $2 \times 2/2$ & $5\times 5 \times 192$ & -\\
\hline
fc1 & - & 512 & 2,457K \\
MFM\_fc1 & - & 256 & - \\
\hline
Total & - & - & 4,095K\\
\hline
\end{tabular}
\label{tab:network_4layer}
\end{table}

Since Network in Network (NIN) \cite{min2013network} can potentially do feature selection between convolution layers and the number of parameters can be reduced by using small convolution kernels like in VGG \cite{simonyan2014very}, we integrate NIN and a small convolution kernel size into the network with MFM. The constructed 9-layer Light CNN contains 5 convolution layers, 4 Network in Network (NIN) layers, Max-Feature-Map layers and 4 max-pooling layers as shown in Table \ref{tab:network_9layer}. The Light CNN-9 contains about 5,556K parameters and 1G FLOPS in total, which is deeper and faster than the Light CNN-4 model.

\begin{table}[h]
\centering
\caption{The architectures of the Light CNN-9 model.}
\begin{tabular}{|c|c|c|c|}
\hline
Type &\begin{tabular}{c}Filter Size\\/Stride, Pad\end{tabular} & Output Size & \#Params \\
\hline
Conv1 & $5 \times 5/1, 2$ & $128\times 128 \times 96$ & 2.4K\\
MFM1 & - & $128\times 128 \times 48$ & - \\
\hline
Pool1 & $2 \times 2/2$ & $64\times 64 \times 48$ & -\\
\hline
Conv2a & $1 \times 1/1$ & $64\times 64 \times 96$ & 4.6K\\
MFM2a & - & $64\times 64 \times 48$ & - \\
Conv2 & $3 \times 3/1, 1$ & $64\times 64 \times 192$ & 165K\\
MFM2 & - & $64\times 64 \times 96$ & - \\
\hline
Pool2 & $2 \times 2/2$ & $32\times 32 \times 96$ & -\\
\hline
Conv3a & $1 \times 1/1$ & $32\times 32 \times 192$ & 18K\\
MFM3a & - & $32\times 32 \times 96$ & - \\
Conv3 & $3 \times 3/1, 1$ & $32\times 32 \times 384$ & 331K\\
MFM3 & - & $32\times 32 \times 192$ & - \\
\hline
Pool3 & $2 \times 2/2$ & $16\times 16 \times 192$ & -\\
\hline
Conv4a & $1 \times 1/1$ & $16\times 16 \times 384$ & 73K\\
MFM4a & - & $16\times 16 \times 192$ & - \\
Conv4 & $3 \times 3/1, 1$ & $16\times 16 \times 256$ & 442K\\
MFM4 & - & $16\times 16 \times 128$ & - \\
\hline
Conv5a & $1 \times 1/1$ & $16\times 16 \times 256$ & 32K\\
MFM5a & - & $16\times 16 \times 128$ & - \\
Conv5 & $3 \times 3/1, 1$ & $16\times 16 \times 256$ & 294K\\
MFM5 & - & $16\times 16 \times 128$ & - \\
\hline
Pool4 & $2 \times 2/2$ & $8\times 8 \times 128$ & -\\
\hline
fc1 & - & 512 & 4,194K \\
MFM\_fc1 & - & 256 & - \\
\hline
Total & - & - & 5,556K\\
\hline
\end{tabular}
\label{tab:network_9layer}
\end{table}

With the development of residual networks \cite{HeZRS16}, the very deep convolution neural networks are widely used and often obtain high performance in various computer vision tasks. We also introduce the idea of residual blocks to Light CNN and design a 29-layer convolution network for face recognition. The residual block contains two $3\times 3$ convolution layers and two MFM operations without batch normalization. There are 12,637K parameters and about 3.9G FLOPS in Light CNN-29. The details of Light CNN-29 are presented in Table \ref{tab:network_29layer}.

\textbf{Note that} there are some differences between the proposed residual block with MFM operations and the original residual block \cite{HeZRS16}. On the one hand, we remove batch normalization from the original residual block. Although batch normalization is efficient to accelerate the convergence of training and avoid overfitting, in practice, batch normalization are domain specific which may be failed when test samples come from differen domains compared with training data. Besides, batch statistics may diminish when the size of training minibatches are small.

On the other hand, we employ the fully connected layer instead of the global average pooling layer on the top. In our training scheme, input images are all aligned, so that each node for high-level feature maps contains both semantic and spatial information which may be damaged by the global average pooling.


\begin{table}[t]
\centering
\caption{The architectures of the Light CNN-29 model.}
\begin{tabular}{|c|c|c|c|}
\hline
Type &\begin{tabular}{c}Filter Size\\/Stride, Pad\end{tabular} & Output Size & \#Params \\
\hline
Conv1 & $5 \times 5/1, 2$ & $128\times 128 \times 96$ & 2.4K\\
MFM1 & - & $128\times 128 \times 48$ & - \\
\hline
Pool1 & $2 \times 2/2$ & $64\times 64 \times 48$ & -\\
\hline
Conv2\_x &$\left[\begin{array}{c}3\times3/1,1\\ 3\times3/1,1\end{array}\right]\times 1$ & $64\times 64 \times 48$ & 82K\\
Conv2a & $1 \times 1/1$ & $64\times 64 \times 96$ & 4.6K\\
MFM2a & - & $64\times 64 \times 48$ & - \\
Conv2 & $3 \times 3/1, 1$ & $64\times 64 \times 192$ & 165K\\
MFM2 & - & $64\times 64 \times 96$ & - \\
\hline
Pool2 & $2 \times 2/2$ & $32\times 32 \times 96$ & -\\
\hline
 Conv3\_x & $\left[\begin{array}{c}3\times3/1,1\\ 3\times3/1,1\end{array}\right]\times 2$ & $32\times 32 \times 96$& 662K\\
Conv3a & $1 \times 1/1$ & $32\times 32 \times 192$ & 18K\\
MFM3a & - & $32\times 32 \times 96$ & - \\
Conv3 & $3 \times 3/1, 1$ & $32\times 32 \times 384$ & 331K\\
MFM3 & - & $32\times 32 \times 192$ & - \\
\hline
Pool3 & $2 \times 2/2$ & $16\times 16 \times 192$ & -\\
\hline
Conv4\_x & $\left[\begin{array}{c}3\times3/1,1\\ 3\times3/1,1\end{array}\right]\times 3$ & $16\times 16 \times 192$& 3981K\\
Conv4a & $1 \times 1/1$ & $16\times 16 \times 384$ & 73K\\
MFM4a & - & $16\times 16 \times 192$ & - \\
Conv4 & $3 \times 3/1, 1$ & $16\times 16 \times 256$ & 442K\\
MFM4 & - & $16\times 16 \times 128$ & - \\
\hline
Conv5\_x & $\left[\begin{array}{c}3\times3/1,1\\ 3\times3/1,1\end{array}\right]\times 4$ & $16\times 16 \times 128$& 2356K\\
Conv5a & $1 \times 1/1$ & $16\times 16 \times 256$ & 32K\\
MFM5a & - & $16\times 16 \times 128$ & - \\
Conv5 & $3 \times 3/1, 1$ & $16\times 16 \times 256$ & 294K\\
MFM5 & - & $16\times 16 \times 128$ & - \\
\hline
Pool4 & $2 \times 2/2$ & $8\times 8 \times 128$ & -\\
\hline
fc1 & - & 512 & 4,194K \\
MFM\_fc1 & - & 256 & - \\
\hline
Total & - & - & 12,637K\\
\hline
\end{tabular}
\label{tab:network_29layer}
\end{table}

\subsection{Semantic Bootstrapping for Noisy Labels}

Bootstrapping, also called ``self-training'', provides a simple and effective approach for sample distribution estimation. Its basic idea is that the forward process on a training sample can be modeled by re-sampling and performing inference from original labeled samples to relabeled ones. It can estimate standard errors and confidence intervals for a complex data distribution and is also appropriate to control the stability of the estimation.

Let $x\in X$ and $t$ denote data and labels, respectively. The CNN based on softmax loss function regresses $x$ onto $t$, and the distribution of its predictions can be represented as a conditional probability $p(t|f(x)), \sum_ip(t_i|f(x))=1$. The maximum probability $p(t_i|f(x))$ determines the most convincing prediction label.
Based on the above theories, we propose a semantic bootstrapping method to sample the training data from the large dataset with massive noisy labels. Firstly, we train the Light CNN-9 model on CASIA-WebFace and fine-tune it on the original noisy labeled MS-Celeb-1M dataset. Second, we employ the trained model to relabel the noisy labeled dataset according to the conditional probabilities $p(t_i|f(x))$. And then we retrain Light CNN-9 on the relabeled dataset. Finally, we further re-sample the original noisy labeled dataset by the retrained model and construct the cleaned MS-Celeb-1M dataset.  The details and discussions for bootstrapping the MS-Celeb-1M dataset are shown in Section \ref{43}.

\section{Experiments}

In this section, we evaluate our Light CNN models on various face recognition tasks. We first introduce the training methodology and databases, and then present the comparison with state-of-the-art face recognition methods, as well as algorithmic analysis and detailed evaluation. Finally, we discuss the effectiveness of the semantic bootstrapping method for selecting training dataset.

\subsection{Training Methodology and Preprocessing}\label{41}

To train the Light CNN, we randomly select one face image from each identity as the validation set and the remaining images as the training set. The open source deep learning framework \textit{Caffe} \cite{jia2014caffe} is employed to implement the CNN model\footnote{https://github.com/AlfredXiangWu/face\_verification\_experiment}.
Dropout is used for fully connected layers and the ratio is set to 0.7.
The momentum is set to 0.9, and the weight decay is set to $5\times10^{-4}$ for convolution layers and a fully-connected layer except the fc2 layer. Obviously, the fc1 fully-connected layer contains the face representation that can be used for face verification. Note that the fc2, which contains large number of parameters with the increasing of identities in the training dataset, are not used for feature extraction.
Thus we increase the weight decay of fc2 layer to $5\times10^{-3}$ to avoid overfitting.
The learning rate is set to $1\times10^{-3}$ initially and reduced to $5\times10^{-5}$ gradually. The parameter initialization for convolutional layers and fully-connected layers is Xavier and Gaussian, respectively.

\begin{figure}[t]
\centering
\subfigure[]{\includegraphics[height=2.5cm,width=0.15\textwidth]{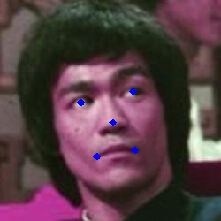}\label{fig:normalization(a)}}
\subfigure[]{\includegraphics[height=2.5cm,width=0.15\textwidth]{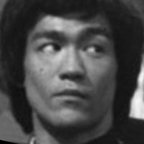}\label{fig:normalization(b)}}
\caption{Face image alignment for training dataset. (a) is the facial points detection results and (b) is the normalized face image.}
\label{fig:normal}
\end{figure}

The CASIA-WebFace and MS-Celeb-1M datasets are used for training in the experiments. To alleviate the influence of large illumination discrepancy, we use gray-scale face images instead of RGB images for training and testing. When training, the face images are aligned to $144\times 144$ by the five landmarks \cite{sun2013deep} (as shown in Fig \ref{fig:normal}) and then randomly cropped to $128\times 128$ as inputs. Besides, each pixel (ranged between $[0, 255]$) is dividing by 255.

\begin{table*}[t]
\centering
\caption{Comparison with other state-of-the-art methods on the LFW and YTF datasets. The unrestricted protocol follows the LFW unrestricted setting and the unsupervised protocol means the model is not trained on LFW in supervised way.}
\begin{tabular}{|c|c|c|c|c|c|c|c|}
\hline
Method & \#Net & Acc on LFW & VR@FAR=0 & Protocol & Rank-1 & DIR@FAR=1\% & Acc on YTF\\
\hline
\hline
DeepFace \cite{taigman2014deepface} & 7 & 97.35 & 46.33 & unrestricted & 64.90 & 44.50 & 91.40\\
Web-Scale \cite{taigman2014web} & 4 & 98.37 & - & unrestricted & 82.50 & 61.90 & - \\
DeepID2+ \cite{DBLP:conf/cvpr/SunWT15}  & 25 & 99.47 &  \textbf{69.36} & unrestricted & \textbf{95.00} & \textbf{80.70} & 93.20\\
WebFace \cite{yi2014learning} & 1 & 97.73 & - & unrestricted & - & - & 90.60\\
FaceNet \cite{schroff2015facenet} & 1 & \textbf{99.63} & - & unrestricted & - & -& 95.10\\
SeetaFace \cite{liu2016viplfacenet} & 1 & 98.62 & -  & unrestricted &  92.79 & 68.13 & - \\
\hline
\hline
VGG \cite{parkhi2015deep} & 1 & 97.27 & 52.40 & unsupervised & 74.10 & 52.01 & 92.80\\
CenterLoss \cite{wen2016discriminative} & 1 & 98.70 & 61.40  & unsupervised & 94.05 & 69.97 & 94.90\\
Light CNN-4  & 1 & 97.97 & 79.20 & unsupervised & 88.79 & 68.03 & 90.72\\
Light CNN-9  & 1 & 98.80 & 94.97 & unsupervised & 93.80 & 84.40 & 93.40\\
Light CNN-29 & 1 & \textbf{99.33} & \textbf{97.50} & unsupervised & \textbf{97.33} & \textbf{93.62} & \textbf{95.54}\\
\hline
\end{tabular}
\label{tab:result}
\end{table*}


\subsection{The Testing Protocols}\label{protocols}

 Four types of face data are used to systematically evaluate the performance of the proposed Light CNN. These databases correspond to large-scale, low-resolution and heterogeneous face recognition (or verification) tasks, respectively. \textbf{Note that we do not re-train or fine-tune the Light CNN model on any testing database.} That is, the training sets of all five testing databases are excluded for model training and fine-tuning.
We directly extract features and compute the similarity of these features measured in cosine similarity.

\textbf{The first type} is the commonly used LFW dataset \cite{huang2007labeled} that contains 13,233 images of 5,749 people. It contains three protocols as follows:
\begin{itemize}\setlength{\itemsep}{1pt}
\item[*] For the standard verification protocol \cite{huang2007labeled}, all face images are divided into 10 folds, each of which contains 600 face pairs without identity overlap.
\item[*] For the probe-gallery identification testing \cite{best2014unconstrained}, there are two new protocols called the close-set and open-set identification. 1) For the close-set task, the gallery set contains 4,249 identities, each with only a single face image, and the probe one contains 3,143 face images belonging to the same set of identities. The performance is measured by Rank-1 identification accuracy. 2) For the open-set task, the gallery set includes 3,143 images of 596 identities. The probe set includes 10,090 images which are constructed by 596 genuine probes and 9,494 impostor ones. The accuracy is evaluated by the Rank-1 Detection and Identification Rate (DIR), which is calculated according to genuine probes matched in Rank-1 at a 1\% False Alarm Rate (FAR) of impostor ones that are not rejected.
\item[*] The Benchmark of Large-scale Unconstrained Face Recognition (BLUFR) \cite{DBLP:conf/icb/LiaoLYL14} is a new benchmark for LFW evaluations, which contains both verification and open-set identification. There are 10-fold experiments, with each fold containing about 156,915 genuine matching and 46,960,863 impostor matching on average for performance evaluation. It is more challenging and generalized compared to LFW.
\end{itemize}

\textbf{The second type} is the collections of video-based face recognition databases containing the YouTube Face (YTF) databse \cite{DBLP:conf/cvpr/WolfHM11}, YouTube Celebrities (YTC) \cite{DBLP:conf/cvpr/KimKPR08} and Celebrity-1000 \cite{DBLP:journals/tcsv/LiuZLY14} that are widely used to evaluate the performance of video-based face recognition methods.
\begin{itemize}\setlength{\itemsep}{1pt}
\item[*] The YTF dataset contains 3,425 videos of 1,595 subjects. Due to low resolution and motion blur, the quality of images in the YTF dataset is worse than in LFW. As for the evaluation protocol, the database is divided into 10 splits, each includes 250 positive pairs and 250 negative ones. As in \cite{parkhi2015deep,schroff2015facenet}, we randomly select 100 samples from each video and compute average similarities.
\item[*] The YTC dataset is composed of 1,910 videos from 47 subjects with a high compression rate and large appearance variations. Following the standard evaluation protocols, the YTC testing set is divided into five-fold cross validation. Each fold contains 423 videos, where the gallery set contains 141 videos and the others are considered as the probe set.
\item[*] The Celebrity-1000 dataset contains 159,726 video sequences from 1000 subjects covering various resolutions, illuminations and poses. There are two types of protocols: close-set and open-set. For the close-set protocol, the training and testing subsets contain the same identities and they are divided into four scales: 100, 200, 500 and 1000 for probe-gallery identification. For the open-set protocol, the generic training set contains 200 subjects and the remaining 800 subjects are used in testing stage. The probe and gallery set are used in testing stage and they are further divided into four scale: 100, 200, 400 and 800.
\end{itemize}

\textbf{The third type} is the very challenging large-scale databases, including the MegaFace \cite{kemelmacher2016megaface}, IJB-A \cite{klare2015pushing} and IJB-B \cite{whitelam2017iarpa} datasets.
\begin{itemize}\setlength{\itemsep}{1pt}
\item[*] MegaFace aims at the evaluation of face recognition algorithms at million-scale. It includes probe and gallery set. The probe set is FaceScrub \cite{DBLP:conf/icip/NgW14}, which contains 100K images of 530 identities, and the gallery set consists of about 1 million images from 690K different subjects.
\item[*] The IJB-A dataset contains 5,712 images and 2,085 videos from 500 identities. All the images and videos are evaluated with two standard protocols, namely, 1:1 face verification and 1:N face identification.
\item[*] The IJB-B dataset is an extension of IJB-A, which contains 11,754 images and 7,011 videos from 1,845 identities. In this paper, we evaluate the Light CNN models on the mixed media (frames and stills) 1:1 verification protocol and open set 1:N protocol using mixed media (frames, stills) as probe.
\end{itemize}

\textbf{The fourth type} is the cross-domain databases, including CACD-VS \cite{DBLP:journals/tmm/ChenCH15}, Multi-PIE \cite{DBLP:journals/ivc/GrossMCKB10} and the CASIA NIR-VIS 2.0 database \cite{DBLP:conf/cvpr/LiYLL13}.
\begin{itemize}\setlength{\itemsep}{1pt}
\item[*] The CACD-VS dataset \cite{DBLP:journals/tmm/ChenCH15} contains large variations in aging. It includes 4,000 image pairs (2000 positive pairs and 2000 negative pairs) by collecting celebrity images on Internet.
\item[*] The Multi-PIE dataset contains 754,204 images of 337 identities under pose, illumination and expression variations in controlled environment.  Following the setting in \cite{DBLP:conf/nips/ZhuLWT14}, 337 subjects with neutral expression, nine pose within $\pm60^{\circ}$ and 20 illuminations are used. The first 200 subjects are for training and the rest 137 subjects are for testing. We do not re-train or finetune Light CNN models on Multi-PIE. Therefore, for a fair comparison, we only evaluate the Light CNN models on the rest 137 identities.
\item[*] The CASIA NIR-VIS 2.0 database for heterogeneous face recognition consists of face images of different modalities. We follow the standard protocol in View 2. There are 10-fold experiments and each fold contains 358 subjects in the testing set. For testing, the gallery set contains only 358 VIS images for each subjects and the probe set consists of 6,000 NIR images from the same 358 subjects.
\end{itemize}

\begin{table}
\centering
\caption{The performance on LFW BLUFR protocols.}
\begin{tabular}{|c|c|c|}
\hline
Method & FAR=0.1\% & DIR@FAR=$1\%$\\
\hline
HighDimLBP \cite{DBLP:conf/icb/LiaoLYL14} & 41.66 & 18.07\\
WebFace \cite{yi2014learning} & 80.26 & 28.90\\
CenterLoss \cite{wen2016discriminative}  & 93.35 & 67.86 \\
\hline
Light CNN-4 & 87.21& 60.24\\
Light CNN-9 & 97.45 & 84.89\\
Light CNN-29 & \textbf{98.88}& \textbf{92.29}\\
\hline
\end{tabular}
\label{tab:blufr}
\end{table}

\subsection{Method Comparison}\label{method_comparison}
\begin{table*}[t]\scriptsize
\caption{The performance on Video-based Face Recognition Databases.}
\begin{minipage}{0.69\textwidth}
\centering
\subtable[Comparison of Rank-1 accuracy (\%) with other state-of-the-art methods on the Celebrity-1000 dataset.]{
\begin{tabular}{|c|c|c|c|c|c|c|c|c|}
\hline
 \multirow{2}{*}{Method} &
 \multicolumn{4}{c|}{Close-Set} &
 \multicolumn{4}{c|}{Open-Set}\\
 \cline{2-9}
   & 100 & 200 & 500 & 1000 & 100 & 200 & 400 & 800\\
\hline\hline
MTJSR \cite{DBLP:journals/tip/YuanLY12} & 50.60 & 40.80 & 35.46 & 30.04 & 46.12 & 39.84 & 37.51 & 33.50\\
DELM \cite{DBLP:journals/corr/UzairSGM15} & 49.80 & 45.21 & 38.88 & 28.83 & - & - & - & -\\
Eigen-RER \cite{DBLP:conf/accv/LiHSLB14} & 50.60 & 45.02 & 39.97 & 31.94 & 51.55 & 46.15 & 42.33 & 35.90\\
GoogleNet+AvePool \cite{yang2017neural} & 84.46 & 78.93 & 77.68 & 73.41 & 84.11 & 79.09 & 78.40 & 75.12\\
GoogleNet+NAN \cite{yang2017neural} & 90.44 & 83.33 & 82.27 & 77.17 & 88.76 & 85.21 & 82.74 & 79.87\\
 \hline\hline
Light CNN-4 & 79.68 & 71.48 & 67.95 & 63.19 & 77.04 & 70.50 & 70.38 & 64.61\\
Light CNN-9 & 81.27 & 74.37 & 72.96 & 67.71 & 77.82 & 75.84 & 74.54 & 68.90\\
Light CNN-29 & \textbf{88.54} & \textbf{81.70} & \textbf{79.62} & \textbf{76.31} & \textbf{85.99} & \textbf{82.38} & \textbf{81.32} & \textbf{77.31}\\
\hline
\end{tabular}
\label{results on C1000}
}
\end{minipage}
\begin{minipage}{0.3\textwidth}
\centering
\subtable[Average Rank-1 Accuracy on YTC dataset.]{
\begin{tabular}{|c|c|}
\hline
Method & Rank-1 Accuracy (\%)\\
\hline
LMKML \cite{DBLP:conf/iccv/LuWM13} & 78.20\\
MMDML \cite{DBLP:conf/cvpr/LuWDMZ15} & 78.50\\
MSSRC \cite{DBLP:conf/cvpr/OrtizWS13} & 80.75\\
SFSR \cite{DBLP:conf/aaai/ZhangHCST16} & 85.74\\
RRNN \cite{DBLP:journals/corr/LiZC16} & 86.60\\
CRG \cite{DBLP:conf/acpr/ChenJTL15} & 86.70\\
VGG \cite{parkhi2015deep} & 93.62 \\
\hline
Light CNN-4 & 88.09\\
Light CNN-9 & 91.56\\
Light CNN-29 & \textbf{94.18}\\
\hline
\end{tabular}
\label{YTC}
}
\end{minipage}
\end{table*}

\begin{table*}\scriptsize
\centering
\caption{The performance on Large-Scale Face Recognition Databases.}
\begin{minipage}[h]{.43\textwidth}
\centering
\subtable[MegaFace performance comparison with other methods on  rank-1 identification accuracy with 1 million distractors and verification TAR for $10^{-6}$ FAR.]{
\begin{tabular}{|c|c|c|}
\hline
Method & Rank-1 & VR@FAR=$10^{-6}$\\
\hline
NTechLAB & 73.300 & 85.081 \\
FaceNet v8 \cite{schroff2015facenet} & 70.496 & 86.473 \\
Beijing Faceall Co. & 64.803 & 67.118 \\
3DiVi Company & 33.705 & 36.927 \\
\hline
Barebones\_FR & 59.363 & 59.036 \\
CenterLoss \cite{wen2016discriminative} & 65.234 & 76.510\\
Light CNN-4 & 60.236& 62.341\\
Light CNN-9 & 67.109 & 77.456\\
Light CNN-29 & \textbf{73.749}& \textbf{85.133}\\
\hline
\end{tabular}
\label{tab:MegaFace}
}
\end{minipage}
\begin{minipage}[h]{0.27\textwidth}
\subfigure[The CMC for MegaFace.]{\includegraphics[width=0.85\textwidth]{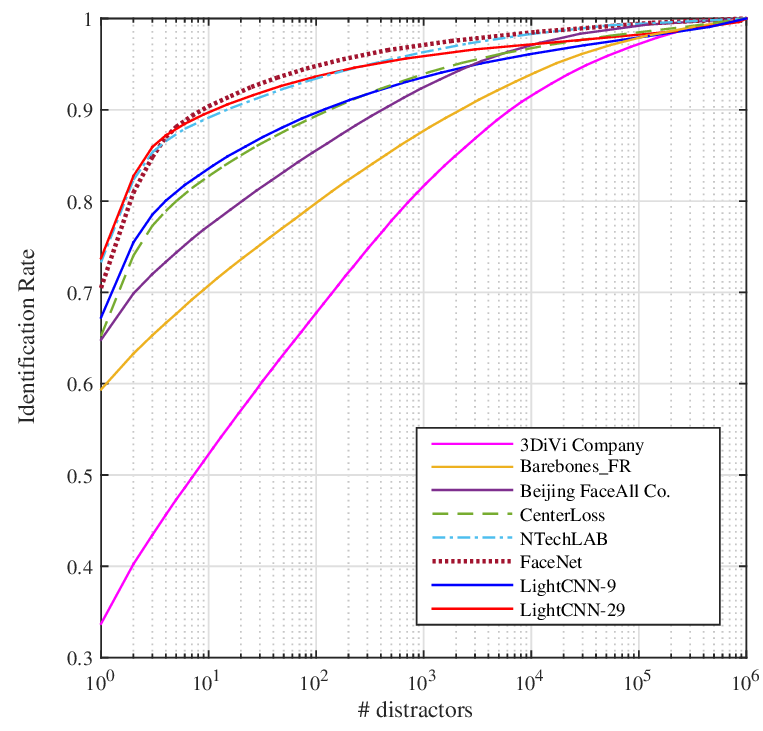}\label{cmc}}
\end{minipage}
\begin{minipage}[h]{0.27\textwidth}
\subfigure[The ROC for MegaFace.]{\includegraphics[width=0.89\textwidth]{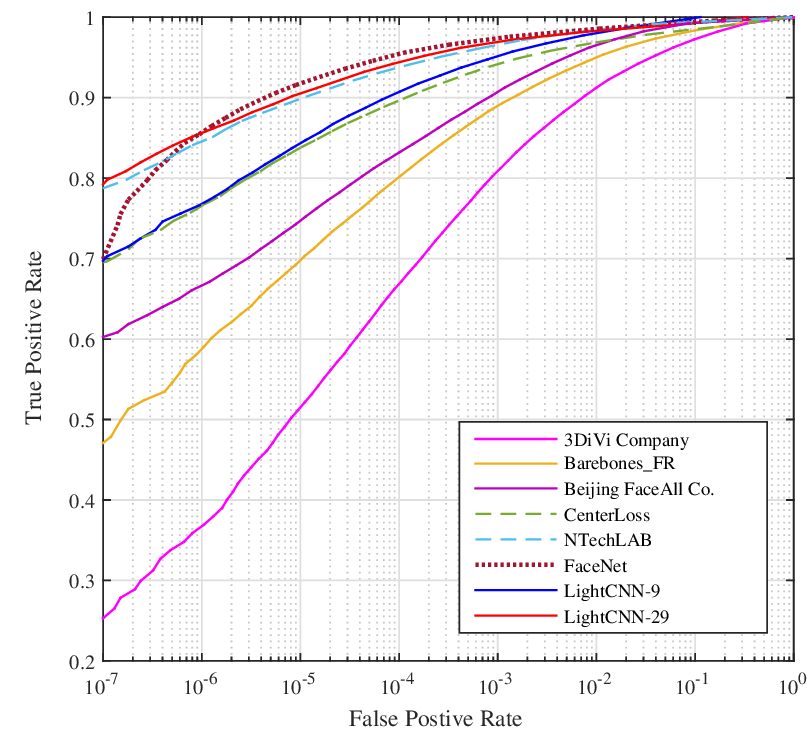}\label{roc}}
\end{minipage}

\subtable[The performance of 1:1 verification and 1:N identification on the IJB-A and IJB-B datasets.]{
\scriptsize
\begin{tabular}{|c|c|c|c|c|c|c|c|c|c|c|}
\hline
\multirow{3}{*}{Method} & \multicolumn{5}{c|}{IJB-A} & \multicolumn{5}{c|}{IJB-B} \\\cline{2-11}
 & \multicolumn{3}{c|}{1:1 Verification} &  \multicolumn{2}{c|}{1:N Identification} & \multicolumn{3}{c|}{1:1 Verification} &  \multicolumn{2}{c|}{1:N Identification}\\\cline{2-11}
& FAR=0.001 & FAR=0.01 & FAR=0.1 & Rank-1 & Rank-5 & FAR=1e-4 & FAR=1e-3 & FAR=1e-2 & Rank-1 & Rank-5\\
\hline
VGGFace \cite{parkhi2015deep} & 62.0$\pm$4.3 & 83.4$\pm$2.1& 95.4$\pm$0.5& 92.5$\pm$0.8 & 97.2$\pm$0.5 & 53.5 & 71.1 & 85.0 & 75.2$\pm$3.8 & 84.3$\pm$3.2\\
Bansalit et al \cite{bansal2017s} & 73.0 & 87.4 & 96.0 & - & -  & -  & -  & - & -  & - \\
Sohn et al \cite{sohn2017unsupervised} & 64.9$\pm$2.2 & 86.4$\pm$0.7& 97.0$\pm$0.1& 89.5$\pm$0.3 & 95.7$\pm$0.2 & -  & -  & - & -  & - \\
Crosswhite et al \cite{crosswhite2017template} & 83.6$\pm$2.7 & 93.9$\pm$1.3& 97.9$\pm$0.4& 92.8$\pm$1.0 & 97.7$\pm$0.4 & -  & -  & - & -  & - \\
NAN \cite{yang2017neural} & 88.1$\pm$1.1 & 94.1$\pm$0.8& 97.8$\pm$0.3& 95.8$\pm$0.5 & 98.0$\pm$0.5 & -  & -  & - & -  & - \\
DA-GAN \cite{zhao2017dual} & 93.0$\pm$0.5 & 97.6$\pm$0.7& 99.1$\pm$0.3& 97.1$\pm$0.7 & 98.9$\pm$0.3 & -  & -  & - & -  & - \\
VGGFace2 \cite{DBLP:journals/corr/abs-1710-08092}& 92.1$\pm$1.4 & 96.8$\pm$0.6& 99.0$\pm$0.2& 98.2$\pm$0.4 & 99.3$\pm$0.2 & 83.1 & 90.8 & 95.6 & 90.2$\pm$3.6 & 94.6$\pm$2.2\\
Whitelam et al \cite{whitelam2017iarpa} & - & - & - & - & - & 54.0 & 70.0 & 84.0 & 79.0 & 85.0\\
\hline
Light CNN-4  & 60.1$\pm$5.3 & 79.4$\pm$3.2& 91.4$\pm$2.1& 88.5$\pm$1.8 & 93.2$\pm$0.7 & 30.5 & 62.4 & 76.5 & 67.5$\pm$4.2 & 78.3$\pm$5.1\\
Light CNN-9 & 83.4$\pm$1.7 & 90.3$\pm$1.0& 95.4$\pm$0.7& 91.6$\pm$0.8 & 95.3$\pm$0.9 & 31.7 & 67.4 & 79.9 & 72.3$\pm$5.8 & 80.4$\pm$4.2\\
Light CNN-29 & \textbf{93.9$\pm$0.9} & \textbf{96.9$\pm$0.4} & \textbf{98.7$\pm$0.1}& \textbf{97.7$\pm$0.3} &\textbf{99.0$\pm$0.1} & \textbf{87.7} & \textbf{92.0} & \textbf{95.3} & \textbf{91.9$\pm$1.5} & \textbf{94.8$\pm$0.9}\\
\hline
\end{tabular}
\label{tab:IJBA}
}
\end{table*}



In this subsection, we train three Light CNN models with MFM 2/1 by the cleaned MS-Celeb-1M dataset. The cleaned MS-Celeb-1M dataset contains 79,077 identities totally about 5,049,824 images that are selected by the proposed semantic bootstrapping method. The details of bootstrapping the dirty MS-Celeb-1M dataset is presented in Section \ref{43}. To certificate the effectiveness of three architectures of the Light CNN, we evaluate our architectures on four different types of face images, including face in the wild, video-based face dataset, large-scale face dataset and cross-domain face dataset. The 256-D deep features are extracted from the output of fully connected layer after MFM operation (MFM\_fc1). The similarity scores are computed by cosine distance and the results of three Light CNN architectures, denoted as Light CNN-4, Light CNN-9 and Light CNN-29, respectively, are shown in tables \ref{tab:result}-VIII.

\subsubsection{Labeled Face in the Wild Database}
On the LFW database (as shown in Table \ref{tab:result}), we evaluate our 4-layer, 9-layer and 29-layer Light CNN models with unsupervised setting, which means our models are not trained or fine-tuned on LFW in a supervised way. The results of our models on the LFW verification protocol are better than those of DeepFace\cite{taigman2014deepface}, DeepID2+ \cite{DBLP:conf/cvpr/SunWT15}, WebFace \cite{yi2014learning}, VGG \cite{parkhi2015deep}, CenterLoss \cite{wen2016discriminative} and SeetaFace \cite{liu2016viplfacenet} for \textbf{a single net}.
Although several business methods have achieved ultimate accuracy on 6000-pairs face verification task, a more practical criterion may be the verification rate at the extremely low false acceptance rate (eg., VR@FAR=0). We achieve \textbf{97.50\%} at VR@FAR=0 for Light CNN-29, while other methods' results are lower than 70\%. Moreover, open-set identification rate at a low false acceptance rate is even more challenging but meaningful in real applications. As shown in Table \ref{tab:result}, Light CNN-29 obtains \textbf{93.62\%} and outperforms DeepFace, DeepID2+, SeetaFace, CenterLoss and VGG. These results suggest that MFM operations are effective for different general CNN architectures and our Light CNNs learn more discriminative embeddings than other CNN methods.

On the BLUFR protocols (as shown in Table. \ref{tab:blufr}), Light CNN obtains \textbf{98.88\%} on TPR@FAR=0.1\% for face verification and  \textbf{92.29\%} on DIR@FAR=1\% for open-set identification, which also obtains better results compared to other state-of-the-art methods.

\begin{table*}
\centering
\caption{The performance on Cross Domain Face Recognition Databases}
\begin{minipage}{0.34\textwidth}
\subtable[Accuracy of different methods on CACD-VS.]{
\begin{tabular}{|c|c|}
\hline
Method & Accuracy (\%) \\
\hline
HD-LBP \cite{chen2013blessing} & 81.60 \\
HFA \cite{DBLP:conf/iccv/GongLLLT13}  & 84.40\\
CARC \cite{DBLP:journals/tmm/ChenCH15}  & 87.60\\
\hline
VGG \cite{parkhi2015deep} & 96.00\\
SeetaFace \cite{liu2016viplfacenet} & 95.50\\
Light CNN-4 & 95.50\\
Light CNN-9 & 97.95\\
Light CNN-29 & \textbf{98.55}\\
\hline
Human, Voting & 94.20 \\
\hline
\end{tabular}
\label{tab:CACD_result}
}
\end{minipage}
\begin{minipage}{0.35\textwidth}
\centering
\subtable[Comparison of Rank-1 accuracy (\%) with other state-of-the-art methods on the Multi-PIE dataset.]{
\begin{tabular}{|c|c|c|c|c|}
\hline
Method & $\pm15^{\circ}$ & $\pm30^{\circ}$ & $\pm45^{\circ}$ & $\pm60^{\circ}$ \\
\hline
Zhu et al. \cite{DBLP:conf/iccv/ZhuLWT13} &  90.7 & 80.7 & 64.1 & 45.9\\
Zhu et al. \cite{DBLP:conf/nips/ZhuLWT14} &  92.8 & 83.7 & 72.9 & 60.1 \\
Kan et al. \cite{DBLP:conf/cvpr/KanSC16} & 100 & \textbf{100} & 90.6 & 85.9  \\
Yin et al. \cite{DBLP:journals/corr/YinL17}& 99.2 & 98.0 & 90.3 & 92.1 \\
Yim et al. \cite{DBLP:conf/cvpr/YimJYCPK15} & 95.0 & 88.5 & 79.9 & 61.9 \\
Tran et al. \cite{LuanCVPR17} & 94.0 & 90.1 & 86.2 & 83.2 \\
\hline
Light CNN-4 & 90.1 & 78.1 & 59.9 & 25.4 \\
Light CNN-9  & 92.1 & 78.2 & 63.6 & 35.7 \\
Light CNN-29 & \textbf{100} & 99.9 & \textbf{99.6} & \textbf{95.0} \\
\hline
\end{tabular}
\label{multipie}
}
\end{minipage}

\begin{minipage}{0.4\textwidth}
\centering
\subfigure[The ROC on CASIA NIR-VIS 2.0 dataset.]{\includegraphics[width=0.70\textwidth]{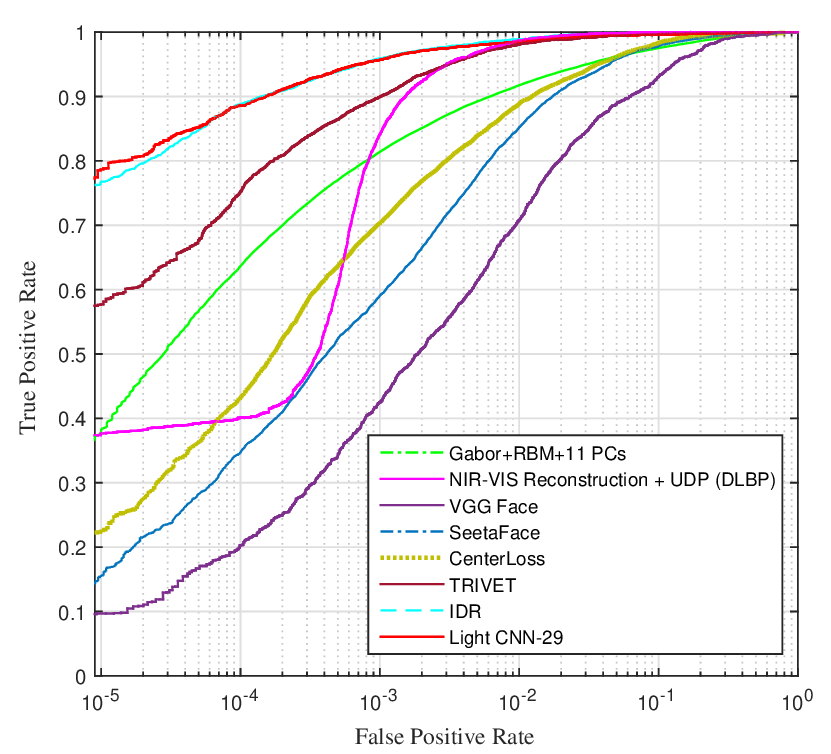}\label{nir_roc}}
\end{minipage}
\begin{minipage}{0.45\textwidth}
\centering
\subtable[Rank-1 accuracy and VR@FAR=0.1\% of different methods on CASIA 2.0 NIR-VIS Face Database.]{
\begin{tabular}{|c|c|c|}
\hline
Method & Rank-1 (\%) & VR@FAR=0.1\% (\%) \\
\hline
Gabor+RBM \cite{DBLP:conf/fgr/YiLL15} & 86.16$\pm$0.98 & 81.29$\pm$1.82\\
DLBP \cite{DBLP:conf/cvpr/Juefei-XuPS15} & 78.46$\pm$1.67 & 85.80\\
TRIVET \cite{DBLP:conf/icb/LiuSWT16} & 95.74$\pm$0.52& 91.03$\pm$1.26\\
IDR \cite{DBLP:conf/aaai/RanIDR17} & 95.82$\pm$0.76& 94.03$\pm$1.06\\
\hline
VGG \cite{parkhi2015deep} & 62.09$\pm$1.88& 39.72$\pm$2.85 \\
SeetaFace \cite{liu2016viplfacenet} & 68.03$\pm$1.66& 58.75$\pm$2.26\\
CenterLoss \cite{wen2016discriminative} &87.69$\pm$1.45 & 69.72$\pm$2.07\\
Light CNN-4 & 85.45$\pm$1.65& 70.91$\pm$1.95\\
Light CNN-9 & 91.88$\pm$0.58 & 85.31$\pm$0.95\\
Light CNN-29 & \textbf{96.72$\pm$0.23}& \textbf{94.77$\pm$0.43}\\
\hline
\end{tabular}
\label{tab:NIR_VIS_result}
}
\end{minipage}
\end{table*}

\subsubsection{Video-based Face Recognition Databases}
Due to low resolution and motion blur, the quality of images in YTF is worse than in LFW. The Light CNN-29 obtains \textbf{95.54\%} without fine-tuning on YTF by using a single model, which outperforms other state-of-the-art methods, such as DeepFace, DeepID2+, WebFace, FaceNet, SeetaFace, VGG and CenterLoss.

As is shown in Table \ref{YTC}, we compare the Light CNN with other video-based face recognition methods such as Localized Multi-Kernel Metric Learning (LMKML) \cite{DBLP:conf/iccv/LuWM13}, Multi-Manifold Deep Metric Learning (MMDML) \cite{DBLP:conf/cvpr/LuWDMZ15}, Mean Sequence Sparse Representation-based Classification (MSSRC) \cite{DBLP:conf/cvpr/OrtizWS13},  Simultaneous Feature and Sample Reduction (SFSR) \cite{DBLP:conf/aaai/ZhangHCST16},  Recurrent Regression Neural Network (RRNN) \cite{DBLP:journals/corr/LiZC16},  Covariate-Relation Graph (CRG) \cite{DBLP:conf/acpr/ChenJTL15} and VGG \cite{parkhi2015deep}. Obviously, the Light CNN-29 obtains \textbf{94.18\%} rank-1 accuracy that is superior to other state-of-the-art methods.

We further evaluate the performance of the Light CNN models on Celebrity-100. The main competitors are Multi-task Joint Sparse Representation (MTJSR) \cite{DBLP:journals/tip/YuanLY12}, Eigen Probabilistic Elastic Part (Eigen-PEP) \cite{DBLP:conf/accv/LiHSLB14}, Deep Extreme Learning Machines (DELM) \cite{DBLP:journals/corr/UzairSGM15} and Neural Aggregation Network (NAN) \cite{yang2017neural}.
In Table \ref{results on C1000}, the Light CNN-29 outperforms its competitors such as MTJSR \cite{DBLP:journals/tip/YuanLY12}, DELM \cite{DBLP:journals/corr/UzairSGM15}, Eigen-RER \cite{DBLP:conf/accv/LiHSLB14} and GoogleNet+AvePool \cite{yang2017neural} on both close-set and open-set protocols. The performance of the Light CNN-29 is worse than that of GoogleNet+NAN \cite{yang2017neural}. This is because the Light CNN models are not trained on Celebrity-1000 and we only employ average pooling along each feature dimension for aggregation as described in \cite{yang2017neural}.

\subsubsection{Large-Scale Face Recognition Databases}
On the challenging MegaFace database (as shown in Table \ref{tab:MegaFace}), we compare the Light CNNs against FaceNet \cite{schroff2015facenet}, NTechLAB, CenterLoss \cite{wen2016discriminative}, Beijing Faceall Co., Barebones\_FR and 3DiVi Company. 
The 29-layer Light CNN achieves \textbf{73.75\%} on rank-1 accuracy and \textbf{85.13\%} on VR@FAR=$10^{-6}$ which outperforms Barebones\_FR, CenterLoss, Beijing Faceall Co. and 3DiVi Company. Besides, the Light CNNs obtain satisfying results compared with some commercial face recognition systems such as Google FaceNet and NTechLAB. Note that they achieve better performance than our models due to the large-scale private training dataset (500M for Google and 18M for NTechLAB) and unknown preprocessing techniques. The CMC and ROC curves are shown in Table.\ref{cmc} and Table.\ref{roc}, respectively.

The performance comparison of the Light CNNs with other state-of-the-arts on IJB-A and IJB-B is given in table \ref{tab:IJBA}. It is observed that the proposed Light CNN-29 obtains comparable results with VGGFace2 \cite{DBLP:journals/corr/abs-1710-08092} and DA-GAN \cite{zhao2017dual} on both 1:1 verification and 1:N identification tasks on IJB-A. Note that VGGFace2 employs a much more computational complex model (SENet-50) that is trained on MS-Celeb-1M and finetuned on VGGFace2 dataset and DA-GAN synthesizes profile face images as the training data. The proposed Light CNN-29 is only trained on MS-Celeb-1M and obtains significant improvements especially on TAR@FAR=0.001 (\textbf{93.9\%} vs 93.0\%).

Besides, the IJB-B dataset is an extension of IJB-A, which contains more subjects and images. Under the mixed media verification and identification protocols, the Light CNN-29 significantly improves the performance on TAR@FAR=1e-4 (\textbf{87.7\%} vs 83.1\%) and Rank-1 (\textbf{91.9\%} vs 90.2\%) compared with VGGFace2 \cite{DBLP:journals/corr/abs-1710-08092}.

\subsubsection{Cross Domain Face Recognition Databases}
Table \ref{tab:CACD_result} shows the results on the CACD-VS dataset. The results of our models on CACD-VS is \textbf{98.55\%} and outperform other age-invariant face recognition algorithms \cite{DBLP:journals/tmm/ChenCH15,chen2013blessing,DBLP:conf/iccv/GongLLLT13} and two open source models \cite{parkhi2015deep, liu2016viplfacenet}. This indicates that our Light CNNs are potentially robust for age-variant problems.

We also compare the Light CNNs with multi-view face recognition methods \cite{DBLP:conf/cvpr/KanSC16, DBLP:conf/iccv/ZhuLWT13, DBLP:conf/nips/ZhuLWT14, DBLP:journals/corr/YinL17} and pose-aware face image synthesis \cite{LuanCVPR17, DBLP:conf/cvpr/YimJYCPK15} in Table \ref{multipie}. It is obvious that the Light CNN-29 obtains great performance on Multi-PIE, where the accuracy on $\pm60^{\circ}$ is about \textbf{95.0\%}. Note that all the compared methods are trained on Multi-PIE, while the Light CNN models are trained on MS-Celeb-1M where the imaging condition is quite different from Multi-PIE. An extension of our Light CNN model also achieves higher accuracy by rotating faces~\cite{RHuang:2017, yibohucvpr2018}. The results indicate that the Light CNN framework can efficiently capture the characteristics of different identities and obtain features invariant to pose and illumination for face recognition.

It is interesting to observe that our Light CNN also performs very well on NIR-VIS dataset. It not only outperforms the three CNN methods but also significantly improves state-of-the-art results such as TRIVET  \cite{DBLP:conf/icb/LiuSWT16} and IDR \cite{DBLP:conf/aaai/RanIDR17} that are trained on the CASIA NIR-VIS 2.0 dataset in a supervised way. We improve the best rank-1 accuracy from 95.82\%$\pm$0.76\% to \textbf{96.72\%$\pm$0.23\%} and VR@FAR=0.1\% is further improved from 94.03\%$\pm$1.06\% to \textbf{94.77\%$\pm$0.43\%}. The ROC curves compared with other methods are shown in Table. \ref{nir_roc}. Note that, different from the compared methods, Light CNNs are not fine-tuned on the CASIA NIR-VIS 2.0 dataset. Particularly, our Light CNN is also applicable for Mesh and VIS heterogenous face recognition~\cite{SZhang:2017}.

The improvement made by out Light CNN models may be attributed to the parameter-free activation functions.
Obviously, compared with other CNN methods that are not trained on the cross-modal NIR-VIS dataset, our Light CNNs based on MFM operations rely on a competitive relationship rather than a threshold of ReLU so that it is naturally adaptive to different appearances from different modalities.
All the experiments suggest that the proposed Light CNNs obtain discriminative face representations and have good generalization ability for various face recognition tasks.

\subsection{Network Analysis}

MFM operation plays an important role in our Light CNN models and it would be interesting to take a closer look into it. Hence we give a detail analysis of MFM on the Light CNN-9 model in this subsection.

First, we compare the performance of MFM 2/1 and MFM 3/2 with ReLU, PReLU and ELU on the LFW database. The basic network is Light CNN-9 as is shown in table \ref{tab:network_9layer}. We simply change activation functions and it is obvious that the output channels of ReLU, PReLU and ELU for each layer are $2\times$ compared with MFM 2/1, and $1.5\times$ compared with MFM 3/2. The experimental results of different activation functions are shown in Table~\ref{tab:relu_mfm}. According to \cite{DBLP:conf/bmvc/ZagoruykoK16}, wider models should obtain better performance. However, we observe that MFM 2/1 and MFM 3/2 are generally superior to the other three activation functions.

The reason is that MFM uses a competitive relationship rather than a threshold (or bias) to active a neuron. Since the training and testing sets are from different data sources, MFM has better generalization ability to different sources. Compared with MFM 2/1, MFM 3/2 can further improve performance, indicating that when using MFM, it would be better to keep only a small number of neurons to be inhibited so that more information can be preserved to the next convolution layer. That is, the ratio between input neurons and output neurons should be set to between 1 and 2.

\begin{table}[t]
\centering
\caption{Comparison with different activation functions on LFW verification and identification protocol by the Light CNN-9 model.}
\begin{tabular}{|c|c|c|c|}
\hline
Method & Accuracy &  Rank-1 & DIR@FAR=1\% \\
\hline
ReLU \cite{nair2010rectified} & 98.30 & 88.58  & 67.56  \\
\hline
PReLU \cite{DBLP:conf/iccv/HeZRS15} & 98.17 &  88.30 &   66.30\\
\hline
ELU \cite{DBLP:journals/corr/ClevertUH15} & 97.70 & 84.70  & 62.09 \\
\hline
MFM 2/1 & 98.80   & 93.80 &  84.40\\
\hline
MFM 3/2& \textbf{98.83}   & \textbf{94.97} &  \textbf{88.59}\\
\hline
\end{tabular}
\label{tab:relu_mfm}
\end{table}

Second, as shown in Table \ref{tab:result}-VIII in Section \ref{method_comparison}, it is obvious that all the proposed Light CNN-4, Light CNN-9 and Light CNN-29 models obtain good performance on various face recognition datasets. It is shown that MFM operation is suitable for the different general CNN architectures such as AlexNet, VGG and ResNet.

\begin{figure}[tbp]
\vspace{-2mm}
\centering
\subfigure[The ROC curves on LFW.]{\includegraphics[width=0.24\textwidth]{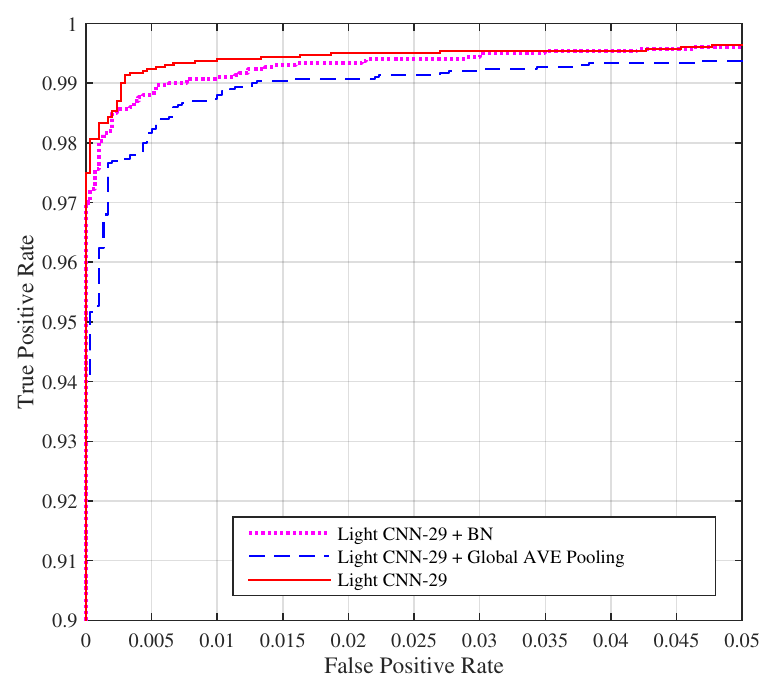}\label{lfw}}
\subfigure[The ROC curves on VAL1.]{\includegraphics[width=0.235\textwidth]{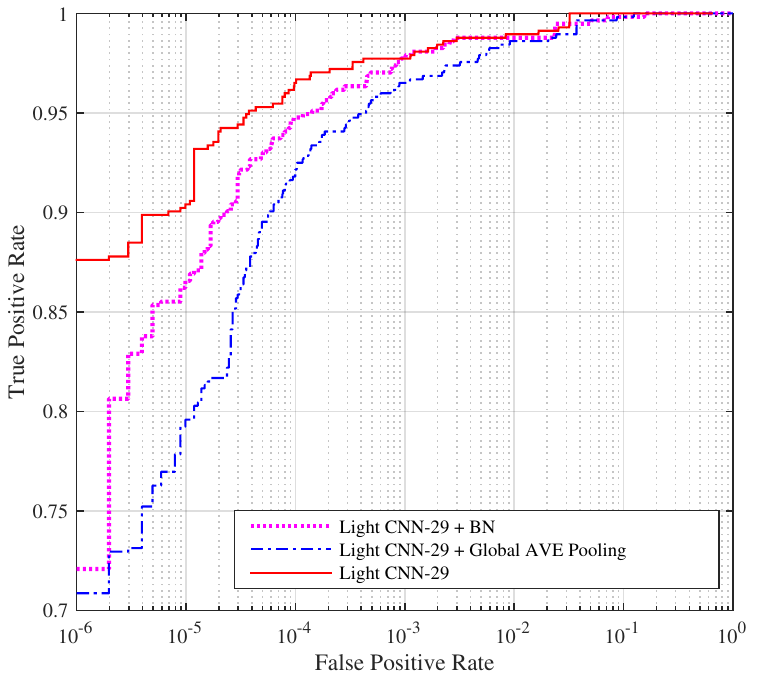}\label{hs}}
\caption{Comparisons on different configurations of Light CNN-29. (a) shows the ROC curves of LFW. (b) shows the ROC curves of VAL1.}
\label{fig_roc}
\end{figure}

Third, we analyze the performance of residual blocks containing MFM operations on two validation datasets. One is LFW which contains 6,000 pairs for evaluations and the other, denoted as VAL1, contains 1,424 face images of 866 identities, that is, 573 positive pairs and 1,012,603 negative pairs. The images in VAL1 are strictly independent from MS-Celeb-1M, as they do not share the same source of data.

In Fig. \ref{fig_roc}, we present the performance of different configurations of Light CNN-29. Obviously, the performance of Light CNN-29 with global average pooling on both LFW and VAL1 is lower than that of Light CNN-29, because the global average pooling does not consider the spatial information of each node in high-level feature map.

In terms of Batch Normalization (BN), as shown in Fig. \ref{lfw}, the model with BN achieves comparative performance as the one without BN. However, Light CNN-29 substantially outperforms the one with BN as shown in Fig. \ref{hs}. This is because most of the images in LFW are portraits of celebrities collected from Internet, which is similar to how MS-Celeb-1M is constructed. However, the images in VAL1 are obtained offline, so that they are strictly independent from the training dataset used.
Obviously, BN diminishes when testing samples are quite independent from the used training datasets, because the means and variances in BN depend on the statistics of the training samples. Based on the above observations, in the Light CNN-29 model, we remove batch normalization and use fully-connected layers instead of global average pooling layers, leading to higher generalization ability.

In addition, computational efficiency is also a critical aspect in comprehensive evaluation of face recognition systems. Very deep CNNs or multi-patch ensembles are common ways to improve recognition accuracy, while they are often time consuming for practical applications.
To verify the computational efficiency of our Light CNNs, we compare our CNN with five widely used models, i.e., FaceNet~\cite{schroff2015facenet}, WebFace~\cite{yi2014learning}, the VGG released model \cite{parkhi2015deep}, the open source SDK SeetaFace \cite{liu2016viplfacenet} and CenterLoss \cite{wen2016discriminative}.

As shown in Table~\ref{tab:speed}, the size of our biggest Light CNN model (Light CNN-29) is $10\times$ smaller than that of the well-known VGG model, while the CPU time is about $5\times$ faster. Compared with the open source face SDK SeetaFace and CenterLoss, Light CNN also performs well in terms of time cost, the number of parameters and feature dimension. Since FaceNet~\cite{schroff2015facenet} and WebFace~\cite{yi2014learning} don't release the models, we report the number of parameters referring to their paper. Obviously, the model of FaceNet~\cite{schroff2015facenet} is approximate 10$\times$ larger than Light CNN-29. In terms of the time cost, the network of FaceNet contains 1.6B FLOPS, while the Light CNN-29 is 3.9G FLOPS. Considering the model of WebFace~\cite{yi2014learning}, although the model size is small, its performance has a large margin compared with Light CNN (shown in Table \ref{tab:result} and Table \ref{tab:blufr}). The results indicate that Light CNN is potentially more suitable and practical on embedding devices and smart phones for real-time applications than its competitors. Particularly, MFM can result in a convolutional neural network with a small parameter space and a small feature dimension. If MFM is well studied and carefully designed, the learned CNN can use a smaller parameter space to achieve better recognition accuracy.

Besides, we implement Light CNNs on MaPU \cite{DBLP:conf/hpca/WangDYLMRWWXWLW16}, a novel architecture which is suitable for data-intensive computing with great power efficiency and sustained computation throughput. The Light CNN-9 for feature extractions are only about 40ms on MaPU, which is implemented by floating-point calculation. It is shown that our light CNNs can be deployed on embedded systems without any precision degradation.

\begin{table}[t]
\centering
\caption{The time cost and the number of parameters of our model compared with VGG, CenterLoss released model and SeetaFace. The speed is tested on a single core i7-4790.}
\begin{tabular}{|c|c|c|c|c|}
\hline
Model & \#Param & \#Dim & Times \\
\hline
FaceNet~\cite{schroff2015facenet} & 140,694K& 128 & - \\
WebFace~\cite{yi2014learning} & 5,015K & 320& -\\
VGG \cite{parkhi2015deep} & 134,251K & 4096&  581ms   \\
SeetaFace \cite{liu2016viplfacenet} & 50,021K & 2048 & 245ms  \\
CenterLoss \cite{wen2016discriminative} & 19,596K & 1024 & 160ms  \\
\hline
Light CNN-4 & 4,095K & 256 & 75ms  \\
Light CNN-9 & 5,556K & 256 & 67ms  \\
Light CNN-29 & 12,637K & 256 & 121ms  \\
\hline
\end{tabular}
\label{tab:speed}
\end{table}

\subsection{Noisy Label Data Bootstrapping}\label{43}

In this subsection, we verify the efficiency of the proposed semantic bootstrapping method on the MS-Celeb-1M database. We select Light CNN-9 for semantic bootstrapping. The testing is performed on LFW. Since LFW only contains 6,000 pairs for evaluations, it is not obvious to justify the effectiveness of methods. We introduce two private datasets in which face images are subject to variations in viewpoint, resolution and illumination.  The first dataset, denoted as VAL1, contains 1,424 face images of 866 identities, consisting of 573 positive pairs and 1,012,603 negative pairs (i.e., totally 1,013,176 pairs). The second dataset contains 675 identities and totally 3,277 images, which is denoted as VAL2. VAL2 contains 2,632,926 pairs that are composed of 4,015 positive and 2,628,911 negative pairs.
All the face images in VAL1 and VAL2 are strictly independent from both CASIA-WebFace and MS-Celeb-1M datasets. Considering highly imbalanced dataset evaluations, we employ Receiver Operator Characteristic (ROC) curves and Precision-Recall (PR) curves to evaluate the performance of the models retrained via bootstrapping. Both VR@FAR for ROC curves and AUC for PR curves are reported.

\begin{table}[t]
\centering
\caption{The performance on \textbf{VAL1}, \textbf{VAL2} and \textbf{LFW} for Light CNN-9 model trained on different databases. It compares the performance of light CNN model trained on CASIA-WebFace, MS-Celeb-1M, MS-Celeb-1M after 1 times bootstrapping(MS-1M-1R) and MS-Celeb-1M after 2 times bootstrapping(MS-1M-2R). The area under Precision-Recall curve(AUC) and verification rate(VR)@fasle acceptance rate(FAR) for different models are presented. }
\begin{tabular}{|c|c|c|c|}
\hline
\textbf{LFW} &ACC (\%) & FAR=1\% (\%) & FAR=0.1\% (\%) \\
\hline
CASIA & 98.13  & 96.73 & 87.13 \\
\hline
MS-Celeb-1M & 98.47  & 98.13 & 94.97  \\
\hline
MS-1M-1R & \textbf{98.80} & 98.43 & 95.43 \\
\hline
MS-1M-2R & \textbf{98.80}  & \textbf{98.60} & \textbf{96.77} \\
\hline
\hline
\textbf{VAL1} &AUC (\%) & FAR=0.1\% (\%) & FAR=0.01\% (\%) \\
\hline
CASIA & 89.72  & 92.50 & 84.82  \\
\hline
MS-Celeb-1M & 92.03  & 94.42 & 88.48  \\
\hline
MS-1M-1R & 94.82  & 96.86 & 92.15  \\
\hline
MS-1M-2R & \textbf{95.34} & \textbf{97.03} & \textbf{93.54}  \\
\hline
\hline
\textbf{VAL2} &AUC (\%) & FAR=0.1\% (\%)& FAR=0.01\% (\%)\\
\hline
CASIA & 62.82  & 62.84 & 44.46 \\
\hline
MS-Celeb-1M & 75.79  & 77.93 & 61.38  \\
\hline
MS-1M-1R & 81.04 & 82.66 & 68.91 \\
\hline
MS-1M-2R & \textbf{82.94}  & \textbf{84.55} & \textbf{71.39} \\
\hline
\end{tabular}
\label{bootstrapping_result}
\end{table}

First, we train a light CNN model on the CASIA-WebFace database that contains about 50K images of 10,575 identities in total. Then, we fine-tune our light CNN on MS-Celeb-1M, initialized by the pre-trained model on CASIA-WebFace. Since MS-Celeb-1M contains 99,891 identities, the fully-connected layer from feature representation (256-d) to identity label, which is treated as the classifier, has a large number of parameters (256*99,891=25,572,096). To alleviate the difficulty of CNN convergence, we firstly set the learning rate of all the convolution layers to 0, so that the softmax loss only contributes to the last fully-connected layer to train a classifier. When it is about to converge, the learning rate of all the convolution layers is set to the same, and then the learning rate is gradually decreased from 1e-3 to 1e-5.

\begin{figure}
\subfigure[The performance trained on MS-Celeb-1M-1R.]{\includegraphics[width=0.24\textwidth]{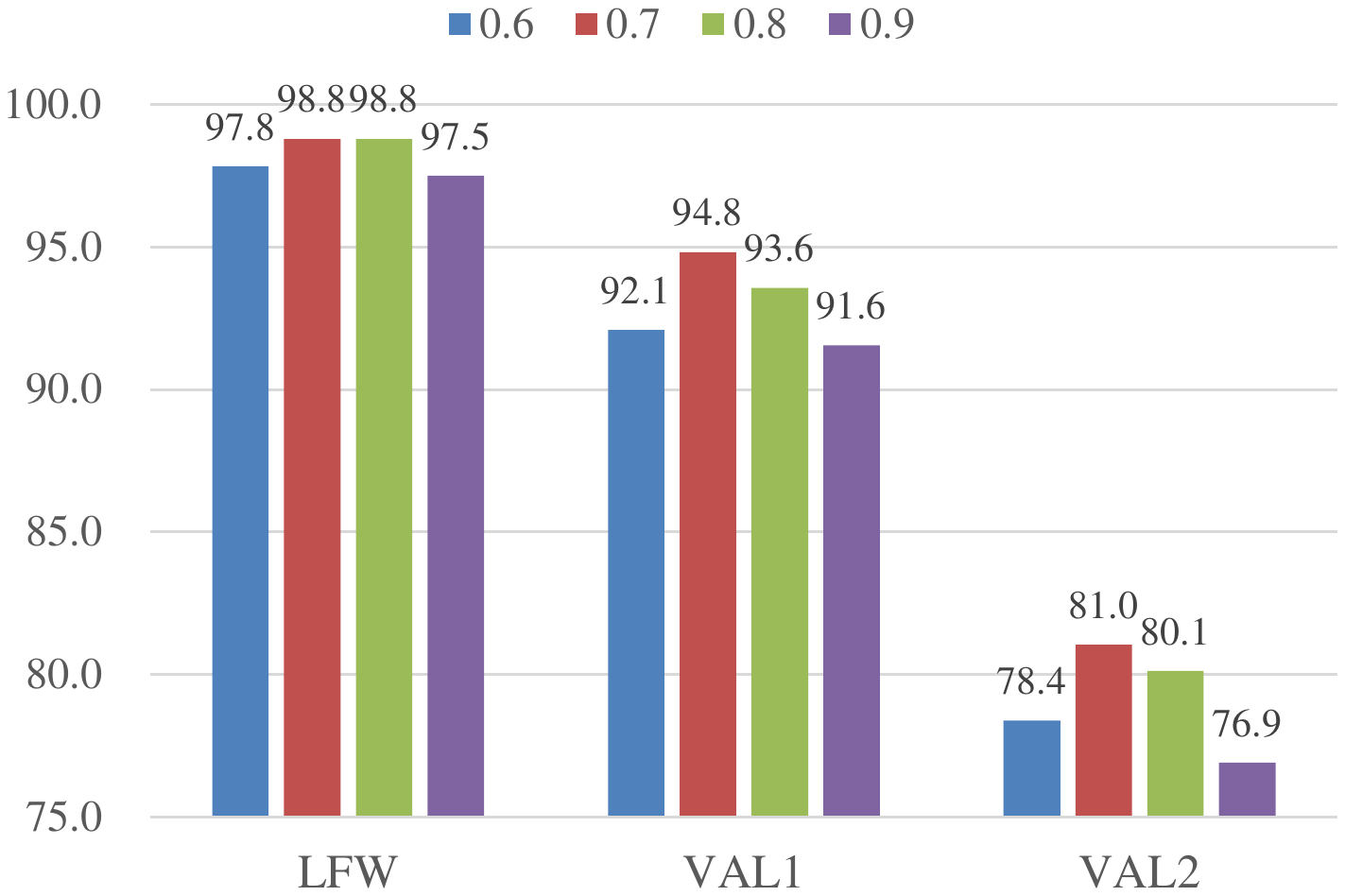}\label{r1}}
\subfigure[The perfomrance trained on MS-Celeb-1M-2R.]{\includegraphics[width=0.24\textwidth]{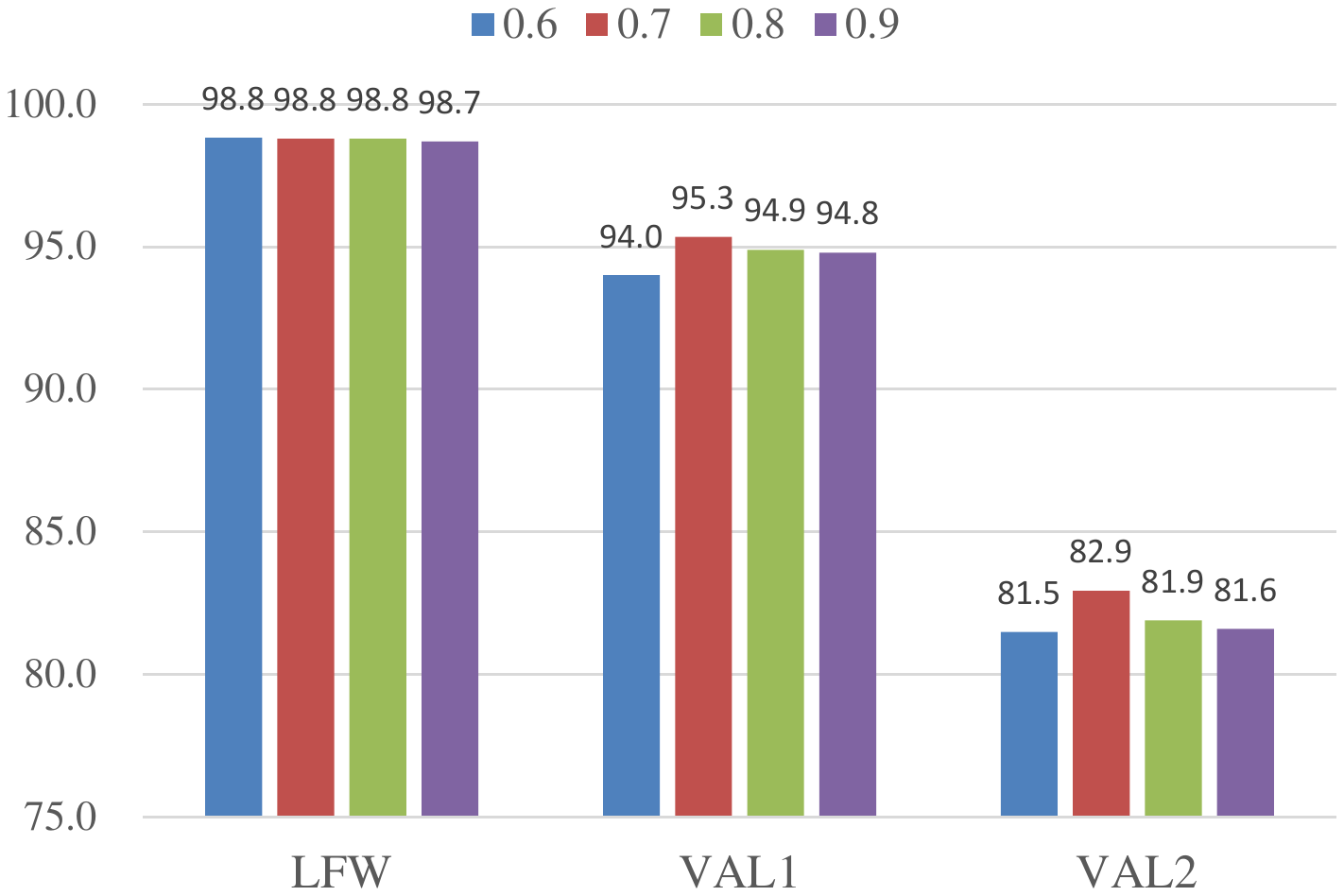}\label{r2}}
\caption{The performance of LFW, VAL1 and VAL2. The models are trained on the cleaned datasets with different threshold settings to sample images.}
\end{figure}

Second, we employ the trained model in the first step to make predictions on the MS-Celeb-1M dataset and obtain the probability $\hat{p_i}$ and label $\hat{t_i}$ for each sample $x_i\in X$. Since the abilities of perceptual consistency can be influenced by the noisy labeled data in the training set, the strict bootstrapping rules are employed to select samples. We accept the re-labeling sample whose prediction label $\hat{t}$ is the same as the ground truth label $t$ and whose probability $\hat{p_i}$ is greater than the threshold $p_0$. As shown in Fig. \ref{r1}, we set $p_0$ to $[0.6, 0.7, 0.8, 0.9]$ to construct four re-labeling datasets. Obviously, the best performance is obtained when $p_0$ is set to 0.7. In this way, the MS-Celeb-1M re-labeling dataset, defined as MS-1M-1R, contains 79,077 identities totally 4,086,798 images.

Third, MS-1M-1R is used to retrain the Light CNN model following the training methodology in Section \ref{41}.
Furthermore, the original noisy labeled MS-Celeb-1M database is re-sampled by the model trained on MS-1M-1R. Assuming that there are few noisy labeled data in MS-1M-1R, we accept the following samples:
1) The prediction $\hat{t}$ is the same as the ground truth label $t$; 2) The prediction $\hat{t}$ is different from the ground truth label $t$, but the probability $p_i$ is greater than the threshold $p_1$. Obviously, the lower threshold $p_1$ is set, the more dangerous we sample an error labeled face image. But if the threshold is set too high, this step is not useful due to less sampled images. According to Fig. \ref{r2}, although LFW obtains nearly the same accuracy for different thresholds, we set $p_1$ to 0.7 due to the best performance on VAL1 and VAL2. The dataset after two times bootstrapping, denoted as MS-1M-2R, contains 5,049,824 images for 79,077 identities.

Finally, we retrain Light CNN-9 on MS-1M-2R. Table \ref{bootstrapping_result} shows experimental results of the CNN models learned on different subsets. We have the following observations: 1) The MS-Celeb-1M database contains massive noisy labels. If the noisy labels are correctly dealt with, the performance on the two testing datasets can be improved. Our semantic bootstrapping method provides a practical way to deal with the noisy labels on the MS-Celeb-1M database. 2) Verification performance benefits from larger datasets. The model trained on the original MS-Celeb-1M database with noisy labels outperforms the model trained on the CASIA-WebFace database in terms of both ROC and AUC. 3)  After two bootstrapping steps, the number of identities drops from 99,891 to 79,077 and performance improvement tends to be smaller. These indicate that our semantic bootstrapping method can obtain a purer training dataset that could in turn result in a light CNN with higher performance.

\section{Conclusions}
In this paper, we have developed a Light CNN framework to learn a robust face representation on noisy labeled dataset. Inspired by neural inhibition and maxout activation, we proposed a Max-Feature-Map operation to obtain a compact and low dimensional face representation. Small kernel sizes of convolution layers, Network in Network layers and Residual Blocks have been implemented to reduce parameter space and improve performance. One advantage of our framework is that it is faster and smaller than other published CNN methods. It extracts one face representation using about 121ms on a single core i7-4790, and it only consists 12,637K parameters for a Light CNN-29 model. Besides, an effective semantic bootstrapping has been proposed to handle the noisy label problem. Experimental results on various face recognition tasks verify that the proposed light CNN framework has potential value for some real-time face recognition systems.

\section*{Acknowledgment}

We would like to thank the associate editor and the reviewers for their valuable comments and advice. This work is funded by State Key Development Program (Grant No. 2016YFB1001001) and the National Natural Science Foundation of China (Grants No. 61622310 and 61427811).

{\small
\bibliographystyle{IEEEtran}
\bibliography{IEEEexample, IEEEtran}
}

\end{document}